# Inducing Alignment Structure with Gated Graph Attention Networks for Sentence Matching

**Peng Cui**, **Le Hu**, and **Yuanchao Liu**
School of Computer Science and Technology, Harbin Institute of Technology, China
`{pcui, lhu, lyc}@insun.hit.edu.cn`

## Abstract

Sentence matching is a fundamental task of natural language processing with various applications. Most recent approaches adopt attention-based neural models to build word- or phrase-level alignment between two sentences. However, these models usually ignore the inherent structure within the sentences and fail to consider various dependency relationships among text units. To address these issues, this paper proposes a graph-based approach for sentence matching. First, we represent a sentence pair as a graph with several carefully design strategies. We then employ a novel gated graph attention network to encode the constructed graph for sentence matching. Experimental results demonstrate that our method substantially achieves state-of-the-art performance on two datasets across tasks of natural language and paraphrase identification. Further discussions show that our model can learn meaningful graph structure, indicating its superiority on improved interpretability.

## 1 Introduction

Sentence matching, which aims to predict the semantic relationship between two sentences, is a fundamental technology for various tasks of natural language processing, such as natural language inference and paraphrase identification. In recent years, deep neural networks have become the dominant method for sentence matching. Existing deep sentence matching approaches can be roughly classified into two categories, which are **representation-based** and **interaction-based**. Representation-based models (Hu et al., 2014; Qiu and Huang 2015; Bowman et al., 2015) focus on sentence-level representation learning. These approaches first encode two sentences into fixed-sized vectors separately, and then compute the matching result on the basis of the two representations. This straightforward paradigm usually has better efficiency because of their sharing encoder parameters and parallel computing. However, the interaction among lower level units of two sentences is ignored (Liu et al., 2019). By contrast, interaction-based approaches adopt matching-aggregation framework (Wang et al., 2017; Chen et al., 2017; Gong et al., 2018; Tay et al., 2018a), which first build the interaction of two sentences at small units (such as words or phrases) and then aggregate these interactive features to predict the final relationship. This framework obtains a remarkable performance gain because it can capture alignment features between two sentences with better granularity (Wang et al., 2017).

Despite the success of recent interaction-based approaches, they still have some limitations. First, most existing works (Chen et al., 2017; Wang et al., 2017; Liu et al., 2019) adopt attention mechanism (Bahdanau et al., 2015) to perform word-by-word comparison between two sentences, which ignores the effect of intra-sentence structure (Liu et al., 2018). Second, some of these approaches suffer from poor interpretability because of the over sophisticated networks or interactive mechanism.
 Language is inherently structured (Liu et al., 2018). An intuitive way is to learn the relationship of two sentences with a graph which has a more complex structure for capturing cross-sentence interactions. However, it is a challenging task to find an effective way to construct and learn the graph for sentence matching task. Efforts have been exerted in various ways. Some studies (Tai et al., 2015; Bowman et al., 2016; Chen et al., 2017) incorporated tree structure into encoding process to learn improved sentence representations. Some other studies (Zhao et al., 2016; Liu et al., 2018) enrich interaction features between two sentences by comparing their subtree structures. However, these approaches solely explore the sentence structure independently and fail to take different edge information (i.e. dependency relations) into consideration. Therefore, how to make the most of graph structure for sentence matching task remains an open question.

To address these issues, this paper proposes a novel graph-based sentence matching approach. Firstly, we explore several carefully designed strategies from different perspectives to model a sentence pair as a semantic graph. Compared with other studies that learn sentence tree or graph structure *independently*, our model works in a more effective way by putting two sentences into one unified graph, where it can simultaneously learn the intra-sentence structure and cross-sentence interaction. Then, we propose a gated graph attention network (G-GAT), which extends vanilla graph attention network (GAT; Veličković et al., 2017) by parameterizing labeled edges as gating vectors to encode dependency relations. We conduct extensive experiments on two datasets across the tasks of natural language inference and paraphrase identification. Experimental results demonstrate the effectiveness and superiority of the proposed approach.

To summarize, our effort provides the following three contributions:

• We propose three carefully designed strategies to model a sentence pair as a semantic graph from different perspectives and conduct systematic exploration on the effects of leveraging graph structure for sentence matching task.

• We proposed a modified GAT model that incorporates gating mechanism to encode various dependency relations, thereby endowing it with better generalization to complex graph with various types of connections.

• The experimental results demonstrate that our approach substantially achieves the state-of-the-art results on two standard datasets. Moreover, analysis of learned graph shows our model induces meaningful alignment structure, indicating its superiority on improved interpretability.

## 2 Related Work

**Neural Sentence Matching Models** Recently, deep neural networks have achieved remarkable results in sentence matching. Early works encoded each sentence separately into a fixed-sized vector and then built a neural classifier upon the two vectors. Recurrent neural networks (Bowman et al., 2015) or convolutional neural networks (Yu et al., 2014; Tan et al., 2016) are widely used as the sentence encoder in this paradigm.

However, these models fail to learn fine-grained interaction features and can hardly model complex relations. Later works, therefore, have adopted matching-aggregation framework that first match sentences at lower level and then aggregates interactive features into final result. Wang et al. (2017) defined several matching strategies to compute alignment from multi-perspectives. Tan et al. (2018) designed multiple attention functions to compare word vectors of two sentences. Gong et al. (2018) used convolutional networks to extract hierarchical information from interaction features. Despite their success, these approaches usually ignore the inherent structure of sentences, which has been proved to plays an important role in learning sentence semantic representation (Tay et al., 2015).

**Graph Structure for Sentence Matching** Due to the complexity of language, some studies focus on leveraging sentence structure to improve the performance of sentence matching. Tay et al. (2015) proposed tree-structured LSTM (Hochreiter and Schmidhuber, 1997) with two variants to encode different parsing trees. Bowman et al. (2016) proposed a modified version with high efficiency for sentence understanding. Chen et al (2017) explored the effects of syntactic information on enriching sentence representation and improving matching performance. These models aimed to encode the intra-sentence structure for learning improved sentence representations. Some other studies learn the relationship of two sentences by comparing their substructures. Among them, Zhao et al. (2016) proposed a subtree-level attention to learn textual entailment. Liu et al., (2018) used a structured attention framework to perform span-level alignment.

Compared to the ways that recent approaches employed to incorporate structure information for sentence matching, we put two sentences into one unified graph. Also, our model takes advantage of the recent advance of graph neural networks (GNNs) and propose a modified GAT which is able to explicitly calculate dependency relations with gating mechanism.

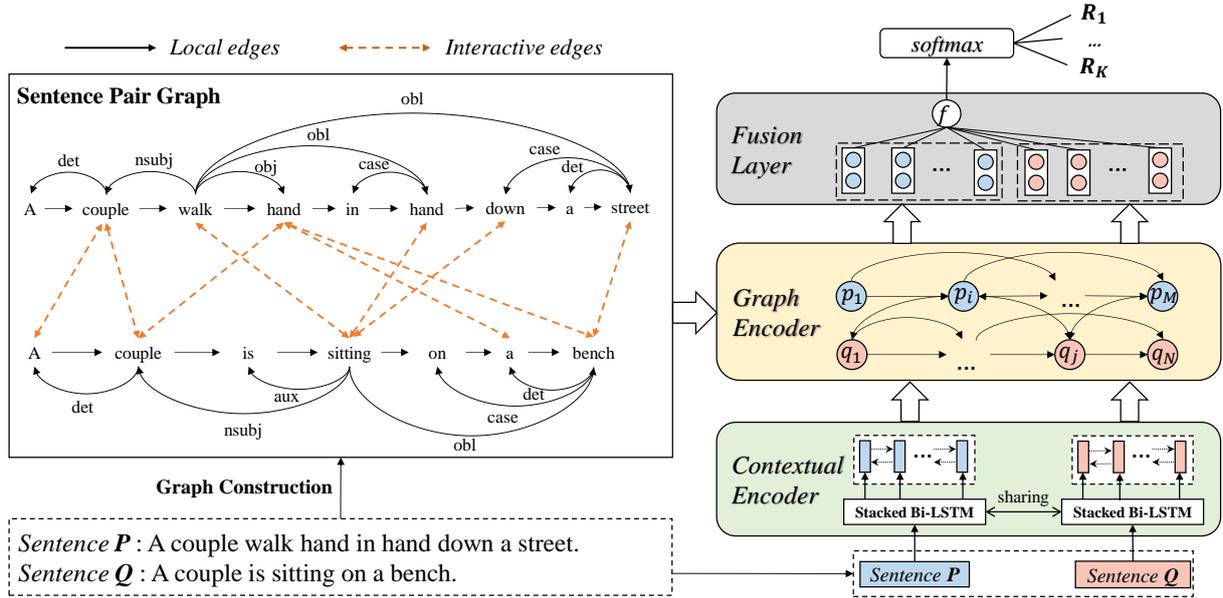

Figure 1. Overall framework of the proposed model (on the right) together with an example of input sentence pair and its constructed semantic graph (on the left).

## 3 Methodology

### 3.1 Overview

This section describes our method, namely, graph attention networks for sentence matching (**Match-Graph**), of which figure 1 provides an illustration. Let $\langle P = \{p_1, p_2 ..., p_M\}, Q = \{q_1, q_2 ..., q_N\}\rangle$ be an arbitrary sentence pair, where $p_i$ and $q_j$ represent the $i$-th word in $P$ and the $j$-th word in $Q$, respectively. The goal of our model is to predict a label $y \in \mathcal{Y}$ representing the relationship between two sentences, where $\mathcal{Y}$ is the predefined relationship set.

Our model consists of three modules playing complementing roles, which are: 1) **contextual encoder**, which uses a stacked recurrent neural network to generate contextual hidden states for each word of two sentences. 2) **graph encoder**, which uses a modified GAT that incorporates gating mechanism to build interaction between textual units of two sentences. 3) **fusion layer**, which aggregates each of the node representation to predict the final relationship. We first describe several strategies for constructing sentence pair graph, then wen explain each part of the model.

### 3.2 Sentence Pair as a Semantic Graph

We start with modeling a sentence pair as a semantic graph. Given a sentence pair $\langle P, Q \rangle$, we use words as graph nodes and create two edge types: 1) *local edges*, which connect nodes within the same sentence, representing the intra-sentence dependency, and 2) *interactive edges*, which connect nodes of different sentences, representing the cross-sentence interaction. By introducing these two edge types, the constructed graph can simultaneously capture inherent structure of both sentences and the interaction features between them. The left side of Figure 1 gives an example of our constructed graph.

Our graph construction can be regarded as a two-steps process. First, we create local edges for two sentences. In particular, we individually perform dependency parsing on the two sentences. Following previous studies (Peng et al., 2017; Quirk and Poon, 2017), we create sequential edges between adjacent words to explicitly infuse the linear dependency. Second, we create interactive edges to merge two sentence graph into one. Let $f_I(p_i, q_j) \to \{0,1\}$ represent whether the edge between words $p_i$ and $q_j$ should be created. We explore three strategies to define construction function $f_I$ from different perspectives. Figure 2 provides an illustration of each strategy.

**Root-based Interaction**  The first strategy to create interactive edges is to simply connect the dependency root of two sentences, which regards each sentence participate as a node in sentence matching. Therefore, this strategy can be consider

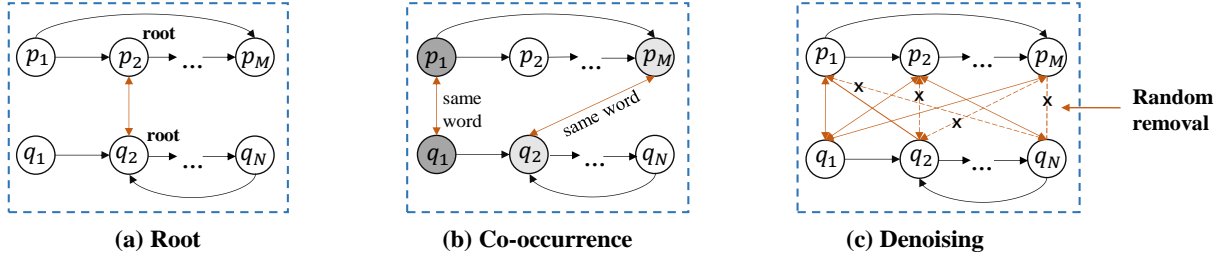

**(a) Root**  **(b) Co-occurrence**  **(c) Denoising**

Figure 2. Strategies of creating interactive edges. (a) direct connecting dependency roots; (b) connecting co-occurrence words; (c) fully connecting each word of two sentences, followed by a denoising step.

as a sentence-level interaction. One potential problem of this intuitive method is that the interaction of other nodes (words) apart from roots may be largely lost because no direct connections exist between them.

$$f_I(p_i, q_j) = \begin{cases} 1, & p_i, q_j = root \\ 0, & otherwise. \end{cases} \quad (1)$$

**Co-occurrence-based Interaction** To gather fine-grained interaction information, we consider an alternative strategy, that is, connecting co-occurrence tokens other than stop words of two sentences. The motivation of this method is that co-occurrence words may serve as strong clues in determining the relationship between two sentences. However, this strategy may struggle with data sparsity because some sentence pairs may have insufficient co-occurrence words.

$$f_I(p_i, q_j) = \begin{cases} 1, & p_i = q_j \\ 0, & otherwise. \end{cases} \quad (2)$$

**Denoising Interaction** The abovementioned proposed methods may experience a common issue where interactive edges are insufficiently constructed. To address this shortcoming, we consider a strategy that can provide maximum interactions, that is, connecting any word of one sentence and any word of the other sentence. However, such fully interaction of two sentences may introduce much noise and slow computational speed because of the large number of redundant interactive edges. Inspired by Vincent et al. (2008), we apply an effective denoising method to randomly remove a proportion of interactive edges.

$$f_I(p_i, q_j) = \begin{cases} 1, & \varepsilon \leq \alpha, \varepsilon \sim U(0,1) \\ 0, & otherwise, \end{cases} \quad (3)$$

Where α is a hyperparameter indicating the keep probability of interactive edges. A larger α represents more interactive edges. In the experiments, we explore a **fully-interaction** strategy (i.e., α=1) to verify the effects of the proposed denoising strategy.

### 3.3 Stacked Contextual Encoder

We have constructed the semantic graph of a given sentence pair $\langle P, Q \rangle$. We then initialize the node representations with a stacked contextual encoder. Specifically, we individually encode two sentences with a Bi-LSTM layer, which has been proven effective for sentence matching (Wang et al., 2017; Duan et al., 2018; Liu et al., 2019), to generate contextual representations of each word.

$$h_{p_i} = \text{Bi-LSTM}(h_{p_{i-1}}), \quad (4)$$

$$h_{q_j} = \text{Bi-LSTM}(h_{q_{j-1}}), \quad (5)$$

where $h_* \in \mathbb{R}^{d_n}$ represents the hidden state of corresponding word and $d_n$ is the predefined hidden size of Bi-LSTM layer. Note that the encoder parameters are shared for the two sentences.

In the experiments, we stack multiple Bi-LSTM layers. The hidden states of last layer $\mathbf{H}^P \in \mathbb{R}^{M \times d_n}$, and $\mathbf{H}^Q \in \mathbb{R}^{N \times d_n}$ are then used to initialized the node representations in the graph encoder, allowing the utilization of contextual information.

### 3.4 Graph Encoder with Relational Gates

Given the constructed graph $\mathcal{G} = (\mathcal{V}, \mathcal{E})$, where nodes set $\mathcal{V}$ correspond to the words, and edges set $\mathcal{E}$ correspond to the local and interactive edges, we employ GAT as the basis of our graph encoder to update node representations.

Let $\mathbf{H^k} = \{h_1^k, h_2^k, \dots, h_{M+N}^k\}$ be the hidden states of input nodes in $k$-th layer, and $\mathbf{H^0} = \mathbf{H}^P \cup \mathbf{H}^Q$ are the outputs from the contextual encoder. The $k$-th GAT layer is designed as:

$$z_{i,j} = LeaklyReLU(W_a[W_e h_i^k; W_e h_j^k]), \quad (5)$$

$$\alpha_{i,j} = \frac{\exp(z_{i,j})}{\sum_{l \in \mathcal{N}_i} \exp(z_{i,l})}, \quad (6)$$

$$h_i^{k+1} = \sum_{j \in \mathcal{N}_i} tanh(\alpha_{i,j} W_c h_j^k), \quad (7)$$

where $\mathcal{N}_i$ stands for neighborhood of the $i$-th node. $\alpha_{i,j}$ stands for the attention weight from node $i$ to $j$. $W_a$, $W_e$, and $W_c$ are model trainable parameters. We use multi-head attention to stabilize the learning process, which can be denoted as

$$h_i^{k+1} = \|_{m=1}^M \sum_{j \in \mathcal{N}_i} tanh(\alpha_{i,j}^m W_c^m h_j^k), \quad (8)$$

The vanilla GAT updates the node representations by aggregating its neighbor nodes. However, this process fails to consider edge information (i.e. different dependency relations). Intuitively, neighbor nodes with different relations should have different influences. To address this issue, we propose a gated graph attention network (G-GAT) to encode various relations. Specifically, we first map each relationship to a low-dimension representation and compute the relational gating vector.

$$g_{i,j} = ReLU(W_g[h_i; h_j; e_{r(i,j)}]), \quad (9)$$

where $e_{r(i,j)} \in \mathbb{R}^{d_r}$ is the vector of relationship between node i and node j with predefined size $d_r$. $W_g$ is trainable parameters. $g_{i,j} \in \mathbb{R}^{d_r}$ is the gating vector, which are then used to control the information flow from nodes $i$ to $j$. Thus, Eq. 7 is modified as follows:

$$h_i^{k+1} = \sum_{j \in \mathcal{N}_i} g_{i,j} \circ tanh(\alpha_{i,j} W_c h_j). \quad (10)$$

By introducing such relational gates, various relationships are parameterized and participate in the graph propagation process. Compared with attention calculation that exerts an alignment score to all the dimension of node vectors, gating mechanism works at better granularity through element-wise operation. Moreover, we choose ReLU activated gates (Xue and Li., 2018) to flexibly control the information flow because it does not have upper bound on important alignment features but strictly declines to zero on noisy information, allowing our model to adaptively learn the complex graph with various relations.

### 3.5 Fusion Layer

We refer to $U_P \in \mathbb{R}^{M \times d_n}$ and $U_Q \in \mathbb{R}^{N \times d_n}$ as the node representations of sentences $P$ and $Q$ after graph encoding, respectively. Intuitively, different nodes contribute unequally for predicting the relationship. Therefore, instead of using direct average summation, we aggregate these vectors with self-attention mechanism following Lin et al. (2017):

$$S_P = \sum_{i=1}^M \alpha_i^P U_P^{(i)}, \quad (11)$$

$$\alpha^P = softxmax(W_1 \tanh(U_P W_P)), \quad (12)$$

$$S_Q = \sum_{j=1}^N \alpha_j^Q U_Q^{(j)}, \quad (13)$$

$$\alpha^Q = softxmax(W_2 \tanh(U_Q W_Q)), \quad (14)$$

where $S_P, S_Q \in \mathbb{R}^{d_n}$ represent the aggregation results of sentence $P$ and sentence $Q$, respectively. $\alpha^P \in \mathbb{R}^M$ and $\alpha^Q \in \mathbb{R}^N$ are the attention weights. $W_P, W_Q, W_1, W_2$ are trainable parameters.

Finally, we combine $s_P$ and $s_Q$ following Mou et al. (2016) and predict the result with two-layer feed-forward network $f_R$.

$$\hat{y} = f_R([s_P; s_Q; s_P - s_Q; s_P \circ s_Q]). \quad (15)$$

In symmetric tasks such as paraphrase identification, we apply a symmetric version of Eq. 12 for better adaptability, i.e., $\hat{y} = f_R([s_P; s_Q; |s_P - s_Q|; s_P \circ s_Q])$.

## 4 Experimental Setup

### 4.1 Datasets

The We evaluated our method on two standard benchmark datasets across two tasks: 1) natural language inference and 2) paraphrase identification. The statistics of two datasets are summarized in Table 1. We briefly introduce the two datasets and their corresponding tasks.

| Dataset | #pairs | | | |R| |
|---|---|---|---|---|
| | Train | Val | Test | |
| SNLI | 550k | 10k | 10k | 3 |
| Quora | 380k | 10k | 10k | 2 |

Table 1. Statistics of two datasets, where |R| denotes the number of classes.

| SNL | | Quora | |
|---|---|---|---|
| **Model** | **Acc. (%)** | **Model** | **Acc. (%)** |
| BiMPM (Wang et al., 2017) | 86.9 | L D C (Wang et al., 2016) | 85.55 |
| ESIM (Chen et al., 2017) | 88.0 | BiMPM (Wang et al., 2017) | 88.17 |
| DIIN (Gong et al., 2018) | 88.0 | DeAttWord (Tomar et al., 2017) | 87.54 |
| MwAN (Tan et al., 2018) | 88.3 | DeAttChar (Tomar et al., 2017) | 88.40 |
| CAFÉ (Tay et al., 2018a) | 88.5 | DIIN (Gong et al., 2018) | 89.06 |
| HIM (Chen et al., 2017) | 88.6 | MwAN (Tan et al., 2018) | 89.12 |
| SAN (Liu et al., 2018) | 88.6 | CSRAN (Tay et al., 2018a) | 89.2 |
| CSRAN (Tay et al., 2018b) | 88.7 | RE2 (Yang et al., 2019) | 89.2 |
| DRCN (Kim et al., 2018) | **88.9** | SAN (Liu et al., 2018) | **89.4** |
| Match-Graph (root) | 87.3±0.10 | Match-Graph (root) | 87.10±0.14 |
| Match-Graph (co-occurrence) | 88.6±0.21 | Match-Graph (co-occurrence) | 89.13±0.26 |
| Match-Graph (fully-interaction) | 88.7±0.13 | Match-Graph (fully-interaction) | 89.17±0.16 |
| Match-Graph (denoising, $\alpha$=0.9) | **88.9±0.05** | Match-Graph (denoising, $\alpha$=0.7) | **89.35±0.08** |

Table 2. Results on SNLI dataset.

**SNLI** (Bowman et al., 2015) is the most widely used benchmark dataset for natural language inference. In natural language inference, the first sentence is called the "premise" and the second is called the "hypothesis". The three predefined labels are "entailment," "neutral," and "contradiction," representing the semantic relations of two sentences. SNLI contains 570k sentence pairs written by humans on the basis of image captioning. We use the standard splits following (Bowman et al., 2015) and use accuracy as the evaluation metric for this dataset.

**Quora Question Pair** is a standard dataset for paraphrase identification, which takes two sentence as input and aims to predict whether one sentence is the paraphrase of the other. Thus, this task can be formulated as a binary classification problem. The Quora dataset contains more than 400k sentence pairs collected from Quora.com. We split the dataset in accordance with previous works (Wang et al., 2017; Yang et al., 2019) and use accuracy as the evaluation metric for this dataset.

### 4.2 Implementation Details

We use AllenNLP (Gardner et al., 2018) as our code framework and implement GNNs with DGL (Wang et al., 2019). The sentences of two datasets are preprocessed by removing punctuation symbols and words appearing less than 10 times. For the graph construction, we use StanfordcoreNLP as dependency parser. The keep probability of interactive edges $\alpha$ is set to 0.9 for the SNLI dataset and 0.7 for Quora Question Pair dataset. For the contextual encoder, we use a 3-layer Bi-LSTM with the same hidden size of 256. For the graph encoder, we set the number of layers to 2 and the number of attention heads to 4 with the hidden size of 64. The size of relation vector is set to 128. Word embeddings are initialized with word2vec (Mikov et al., 2013) of 300D and fine-tuned in training process. All other parameters are randomly initialized. During the training, we use Adam optimizer (Kingma and Ba, 2014) with learning rate set to $5e-4$. We train our model for 300 epochs with batch size of 64. All hyperparameters are selected through grid search on the validation set with accuracy as the metric. We report the average accuracy and the standard deviation over 10 runs.

## 5 Results and Analysis

### 5.1 Main Results

We compare our approach with state-of-the-art models on two datasets. For the sake of fairness, we do not compare it with models equipped with pre-trained language models. Considering our approach is orthogonal to pre-trained encoder, we believe it can be further boosted by integrating them, which we reserve for future work.

Tables 2 presents the accuracy results on SNLI and Quora datasets, respectively. The first sections report the state-of-the-art public approaches, and the second sections report the results of our model with different graph construction strategies. Results that are not significantly distinguished from the best approaches are marked in bold. From the results, we can make the following conclusions.

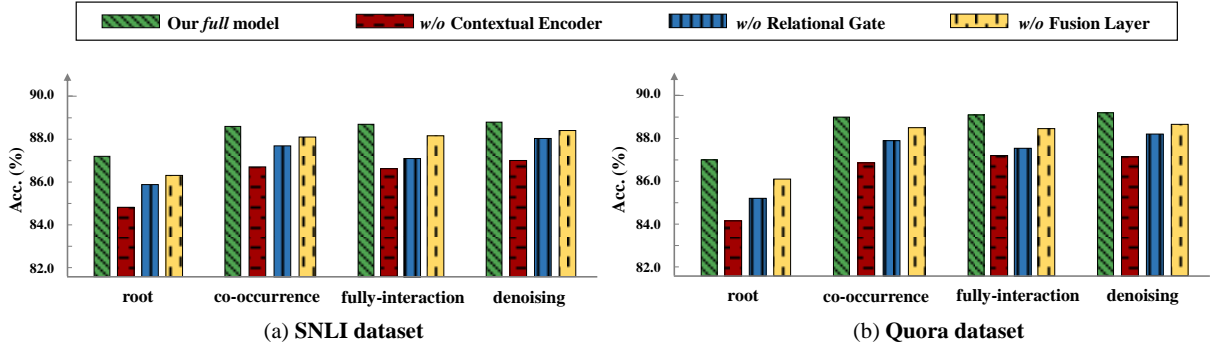

Figure 3. Comparison results of our full model and three variants on the two datasets.

1) Denoising strategy outperforms the fully-interaction strategy and achieves the best results among four versions of our method, demonstrating its effectiveness. 2) Root-based interaction strategy performs worst, proving the necessity of fine-grained interactions. 3) The best version of our model, that is, denoising strategy, obtains results on par with the state-of-the-art models (DRCN on SNLI and SAN on Quora) and outperforms other strong baselines on both datasets. Besides, our model works in a more interpretable way by inducing meaningful graph structure, which will be further discussed in Section 5.5.

### 5.2 . Ablation Study

To analyze the relative contributions of different modules on matching sentences, we compare our full model with three ablated variants: 1) w/o contextual encoder, which removes stacked Bi-LSTM layer and initialize the node vectors with corresponding word embedding. 2) w/o relational gates, which removes gating mechanism of graph encoder. In this variant, the graph encoder degenerates to the vanilla GATs. 3) w/o fusion layer, which aggregates node representations with average pooling instead of self-attention mechanism.

Figure 3 presents the results of different variants on both datasets, from which we can make the following observations. 1) Our full model outperforms all variants on both datasets, which proves that the three modules play complementing roles in the sentence matching process, and combining them can help our model achieve the best performance. 2) When removing contextual encoder, i.e. the stacked Bi-LSTM layer, the performance of each strategy declines dramatically, indicating that contextual information is particular important for sentence matching. This observation suggests that integrating pre-trained contextual encoder into our model may provide our approach additional performance gain. 3) Relational gates substantially boost the performance, particularly for fully-interaction strategy, indicating they can improve the generalization capacity of our model for complex graph structure, which is consistent with the aforementioned analysis.

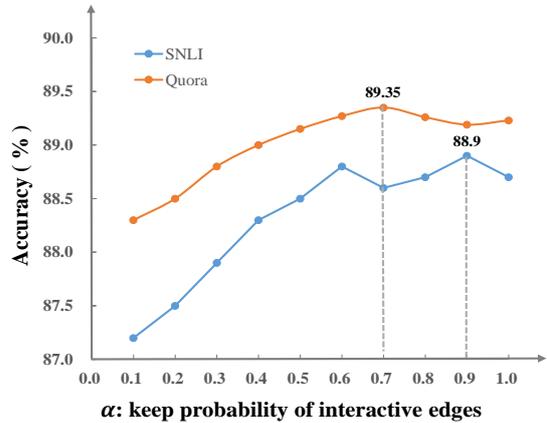

Figure 4. Relationships between hyperparameter α (x-axis) and accuracy (y-axis) on the two datasets.

### 5.3 Effects of Denoising Strategy

As discussed in Section 3.1, hyperparameter α plays a key role in denoising strategy by determining the keep number of interactive edges. Therefore, we conduct experiments to quantitatively explore the influence of α. Figure 3 presents the accuracy under different α on the two datasets.

We can see from figure 4 that two subgraphs show a similar trend. Specifically, the accuracy intensively increases when α is tuned from 0.1 to 0.6, indicating that some amount of interactive edges are necessary for learning the relation of two

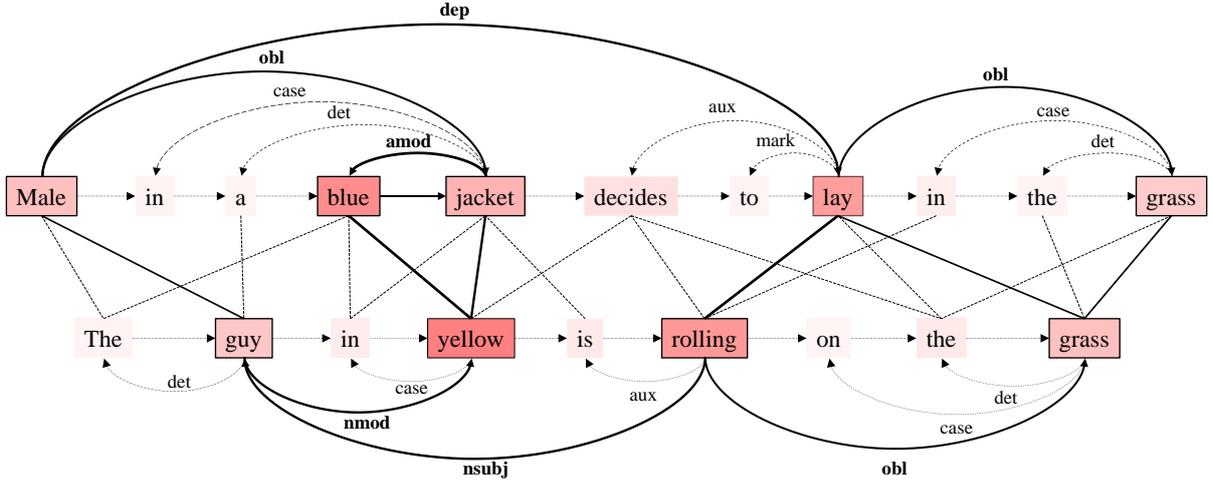

Figure 5. An example of visualized attention distribution on learned semantic graph, with important nodes colored darker and important edges bold. For simplicity, we omit the interactive edges with ex-tremely low weights.

Sentences. However, the accuracy experiences fluctuation when α > 0.6, and the fully-interaction (i.e. α = 1) does not make the model achieve the highest accuracy on either dataset, which demonstrates that excessive edges may introduce noisy information and make the model tend to overfit. This result reflects the influence of the number of interactive edges on the performance to some extent, which provides an instruction for the design of graph construction strategy.

### 5.4 Analysis of Learned Graph Structure

In this subsection, we give a qualitative analysis of the learned semantic structure to briefly understand how our model work. Recall that in fusion layer, we compute the final result with vectors $S_P$ and $S_Q$ (Eq. 10), which are aggregated with the node attentions $\alpha^P$ and $\alpha^Q$ (Eq. 11). Therefore, $\alpha^P$ and $\alpha^Q$ can be considered as the contribution of each node to the result, and the importance of edges can be defined as the weight sum of its head node and tail node.

In Figure 5, we visualize the semantic graph attention distribution of a real data, where the premise sentence is "*Male in a blue jacket decides to lay in the grass*" and the hypothesis sentence is "*The guy in yellow is rolling on the grass*". We can see that the important nodes (colored darker) and edges (solid and bold) form a meaningful subgraph, which indeed presents the alignment structure of two sentences and the inferencing chain of their semantic relationship. For example, the words "*Male*" and "*grass*" have weak alignment with words "*guy*" and "*grass*", respectively, because of their semantic association. However, the words "*blue jacket*" and "*lay*" have strong alignment with words "*yellow*" and "*rolling*", respectively, showing that the model pays more attention on the differences of the two sentences. Consequently, our model correctly predicts the relationship is "contradiction." It is worth noting that long-range important nodes are effectively bridged by syntactic dependency apart from interactive edges, indicating that intra-sentence structure also plays a key role in sentence matching. Moreover, the learned graph can in turn gives an inspiration to the graph construction.

## 6 Conclusion and Future Work

This paper proposes a graph-based sentence matching approach. Firstly, we explore several effective strategies to construct semantic graph for a sentence pair. Then we propose a novel gated graph attention network (G-GAT), which employs gating mechanism to capture various relationships among textual units. Experimental results on two well-studied datasets across tasks of natural language inference and paraphrase recognition demonstrate that the proposed method not only substantially achieves state-of-the-art performance but also can induce interpretable graph structure for sentence matching. In the future, we will explore graph construction strategies by integrating more technologies (e.g., semantic role labeling and constituent parsing) to further enhance the performance.

# References

Samuel R. Bowman, Gabor Angeli, Christopher Potts, and Christopher D. Manning. 2015. A large annotated corpus for learning natural language inference. In *Proceedings of the 2015 Conference on Empirical Methods in Natural Language Processing*, pages 632–642, Lisbon, Portugal. Association for Computational Linguistics.

Samuel R. Bowman, Jon Gauthier, Abhinav Rastogi, Raghav Gupta, Christopher D. Manning, and Christopher Potts. 2016. A fast unified model for parsing and sentence understanding. In *Proceedings of the 54th Annual Meeting of the Association for Computational Linguistics (Volume 1: Long Papers)*, pages 1466–1477, Berlin, Germany.

Dzmitry Bahdanau, Kyunghyun Cho, and Yoshua Bengio. 2015. Neural machine translation by jointly learning to align and translate. In *Proceedings of the 3rd International Conference on Learning Representations*.

Qian Chen, Xiaodan Zhu, Zhen-Hua Ling, Si Wei, Hui Jiang, and Diana Inkpen. 2017. Enhanced LSTM for natural language inference. In *Proceedings of the 55th Annual Meeting of the Association for Computational Linguistics* (Volume 1: Long Papers), pages 1657–1668, Vancouver, Canada. Association for Computational Linguistics.

Chaoqun Duan, Lei Cui, Xinchi Chen, Furu Wei, Conghui Zhu, and Tiejun Zhao. 2018. Attention-fused deep matching network for natural language inference. In *IJCAI*, pages 4033–4040.

Matt Gardner, Joel Grus, Mark Neumann, Oyvind Tafjord, Pradeep Dasigi, Nelson F. Liu, Matthew Peters, Michael Schmitz, and Luke Zettlemoyer. 2018. AllenNLP: A deep semantic natural language processing platform. In *Proceedings of Workshop for NLP Open Source Software (NLP-OSS)*, pages 1– 6, Melbourne, Australia. Association for Computational Linguistics.

Yichen Gong, Heng Luo, and Jian Zhang. 2018. Natural language inference over interaction space. In *Proceedings of the 6th International Conference on Learning Representations*.

Baotian Hu, Zhengdong Lu, Hang Li, and Qingcai Chen. 2014. Convolutional neural network architectures for matching natural language sentences. In *Advances in Neural Information Processing Systems*

Sepp Hochreiter and Jrgen Schmidhuber. 1997. Long short-term memory. *Neural Computation*.

Seonhoon Kim, Jin-Hyuk Hong, Inho Kang, and Nojun Kwak. 2018. Semantic sentence matching with densely-connected recurrent and co-attentive information. *arXiv preprint arXiv:1805.11360. Version 2*.

Diederik Kingma and Jimmy Ba. 2014. Adam: A method for stochastic optimization. In *Proceedings of the 2nd International Conference on Learning Representations*.

Zhouhan Lin, Minwei Feng, Cicero Nogueira dos Santos, Mo Yu, Bing Xiang, Bowen Zhou and Yoshua Bengio. 2017. A Structured Self-attentive Sentence Embedding. *arXiv preprint arXiv:1703.03130*.

Xiaodong Liu, Kevin Duh and Jianfeng Gao. 2019. Stochastic Answer Networks for Natural Language Inference. *arXiv preprint arXiv:1804.07888*.

Yang Liu, Matt Gardner, and Mirella Lapata. 2018. Structured Alignment Networks for Matching Sentences. In *Proceedings of the 2018 Conference on Empirical Methods in Natural Language Processing*, pages 1554–1564 Brussels, Belgium.

Mingtong Liu, Yujie Zhang, Jinan Xu, Yufeng Chen. 2019. Original Semantics-Oriented Attention and Deep Fusion Network for Sentence Matching. In *Proceedings of the 2019 Conference on Empirical Methods in Natural Language Processing and the 9th International Joint Conference on Natural Language Processing*, pages 2652–2661, Hong Kong, China.

Lili Mou, Rui Men, Ge Li, Yan Xu, Lu Zhang, Rui Yan, and Zhi Jin. 2016. Natural language inference by tree-based convolution and heuristic matching. In *Proceedings of the 54th Annual Meeting of the Association for Computational Linguistics (Volume 2: Short Papers)*, pages 130–136, Berlin, Germany.

Tomas Mikolov, Wen-tau Yih, and Geoffrey Zweig. 2013. Linguistic regularities in continuous space word representations. In *Proceedings of the 2013 Conference of the North American Chapter of the Association for Computational Linguistics: Human Language Technologies*, pages 746–751, Atlanta, Georgia.

Nanyun Peng, HoifungPoon, Chris Quirk, KristinaToutanova, and Wen-tauYih. 2017. In *Transactions of the Association for Computational Linguistics*, vol. 5, pp. 101–115, 2017.

Xipeng Qiu and Xuanjing Huang. 2015. Convolutional neural tensor network architecture for community-based question answering. In *Proceedings of International Joint Conference on Artificial Intelligence*.

Chris Quirk and Hoifung Poon. 2017. Distant supervision for relation extraction beyond the sentence boundary. In *Proceedings of the Fifteenth Conference on European chapter of the Association for Computational Linguistics*.

Gaurav Singh Tomar, Thyago Duque, Oscar Tackstrom, Jakob Uszkoreit, and Dipanjan Das. 2017. Neural paraphrase identification of questions with noisy pretraining. In *Proceedings of the First Workshop on Subword and Character Level Models in NLP*, pages 142–147, Copenhagen, Denmark.

Yi Tay, Anh Tuan Luu, and Siu Cheung Hui. 2018a. Compare, compress and propagate: Enhancing neural architectures with alignment factorization for natural language inference. In *Proceedings of the 2018 Conference on Empirical Methods in Natural Language Processing*, pages 1565–1575, Brussels, Belgium. Association for Computational Linguistics.

Yi Tay, Anh Tuan Luu, and Siu Cheung Hui. 2018b. Co-stack residual affinity networks with multi-level attention refinement for matching text sequences. In *Proceedings of the 2018 Conference on Empirical Methods in Natural Language Processing*, pages 4492–4502, Brussels, Belgium.

Kai Sheng Tai, Richard Socher, and Christopher D. Manning. 2015. Improved semantic representations from tree-structured long short-term memory networks. In *Proceedings of the 53rd Annual Meeting of the Association for Computational Linguistics and the 7th International Joint Conference on Natural Language Processing (Volume 1: Long Papers)*, pages 1556–1566, Beijing, China.

Chuanqi Tan, Furu Wei, Wenhui Wang, Weifeng Lv, and Ming Zhou. 2018. Multiway attention networks for modeling sentence pairs. In *Proceedings of the Twenty-Seventh International Joint Conference on Artificial Intelligence*, IJCAI-18, pages 4411–4417. International Joint Conferences on Artificial Intelligence Organization.

Ming Tan, Cicero dos Santos, Bing Xiang, and Bowen Zhou. 2016. Improved representation learning for question answer matching. In *Proceedings of the 54th Annual Meeting of the Association for Computational Linguistics (Volume 1: Long Papers)*, pages 464–473, Berlin, Germany. Association for Computational Linguistics.

Petar Veličković, Guillem Cucurull, Arantxa Casanova, Adriana Romero, Pietro Lio, and Yoshua Bengio. 2017. Graph attention networks. arXiv preprint arXiv:1710.10903.

Pascal Vincent, Hugo Larochelle, Yoshua Bengio, Pierre-Antoine Manzagol. 2008. Extracting and Composing Robust Features with Denoising Autoencoders. In *Proceedings of the 25th International Conference on Machine Learning*, Helsinki, Finland, 2008.

Zhiguo Wang, Wael Hamza, and Radu Florian. 2017. Bilateral multi-perspective matching for natural language sentences. In P*roceedings of the Twenty-Sixth International Joint Conference on Artificial Intelligence, IJCAI-17*, pages 4144–4150.

Zhiguo Wang, Haitao Mi, and Abraham Ittycheriah. 2016b. Sentence similarity learning by lexical decomposition and composition. a*rXiv preprint arXiv:1602.07019.*

Minjie Wang, Lingfan Yu, Da Zheng, Quan Gan, Yu Gai, Zihao Ye, Mufei Li, Jinjing Zhou, Qi Huang, Chao Ma, Ziyue Huang, Qipeng Guo, Hao Zhang, Haibin Lin, Junbo Zhao, Jinyang Li, Alexander J Smola, and Zheng Zhang. 2019. Deep graph library: Towards efficient and scalable deep learning on graphs. *ICLR Workshop on Representation Learning on Graphs and Manifolds.*

Wei Xue and Tao Li. 2018. Aspect Based Sentiment Analysis with Gated Convolutional Networks. In *Proceedings of the 56th Annual Meeting of the Association for Computational Linguistics (Long Papers)*, pages 2514–2523 Melbourne, Australia.

Runqi Yang, Jianhai Zhang, Xing Gao, Feng Ji, Haiqing Chen. 2019. Simple and Effective Text Matching with Richer Alignment Features. In *Proceedings of the 57th Annual Meeting of the Association for Computational Linguistics*, pages 4699–4709.

Lei Yu, Karl Moritz Hermann, Phil Blunsom, and Stephen Pulman. 2014. Deep learning for answer sentence selection. In *NIPS Deep Learning and Representation Learning Workshop, Montreal.*

Kai Zhao, Liang Huang, and Mingbo Ma. 2016. Textual entailment with structured attentions and composition. In *Proceedings of COLING 2016, the 26th International Conference on Computational Linguistics: Technical Papers*, pages 2248–2258, Osaka, Japan.